\pdfoutput=1

\documentclass[11pt]{article}

\usepackage[preprint]{acl}

\usepackage{times}
\usepackage{latexsym}
\usepackage{tabularx}
\usepackage{multirow}
\usepackage{mathtools}
\usepackage{enumitem}
\usepackage{booktabs}
\usepackage{makecell}

\usepackage{listings}
\usepackage{caption}
\usepackage{pifont}
\newtoggle{comment}
\toggletrue{comment}

\usepackage[T1]{fontenc}

\usepackage[utf8]{inputenc}

\usepackage{microtype}

\usepackage{inconsolata}

\usepackage{graphicx}

\title{END: Early Noise Dropping for Efficient and Effective Context Denoising}

\author{Hongye Jin\textsuperscript{$\star$}, Pei Chen, Jingfeng Yang, Zhengyang Wang, Fengran Mo\textsuperscript{$\star$}, Jinghan Zhang\textsuperscript{$\star$}, \\ {\bf Meng Jiang, Yifan Gao, Binxuan Huang, Xinyang Zhang, Zheng Li,} \\ {\bf Tianyi Liu, Huasheng Li, Xia Hu, Bing Yin}}

\begin{document}
\maketitle
\renewcommand{\thefootnote}{$\star$}
\footnotetext{Work done during internship at Amazon. Preprint.}
\renewcommand{\thefootnote}{\arabic{footnote}}
\begin{abstract}
Large Language Models (LLMs) have demonstrated remarkable performance across a wide range of natural language processing tasks. 

However, they are often distracted by irrelevant or noisy context in input sequences that degrades output quality. 
This problem affects both long- and short-context scenarios, such as retrieval-augmented generation, table question-answering, and in-context learning. We reveal that LLMs can implicitly identify whether input sequences contain useful information at early layers, prior to token generation. Leveraging this insight, we introduce Early Noise Dropping (\textsc{END}), a novel approach to mitigate this issue without requiring fine-tuning the LLMs. \textsc{END} segments input sequences into chunks and employs a linear prober on the early layers of LLMs to differentiate between informative and noisy chunks. By discarding noisy chunks early in the process, \textsc{END} preserves critical information, reduces distraction, and lowers computational overhead. Extensive experiments demonstrate that \textsc{END} significantly improves both performance and efficiency across different LLMs on multiple evaluation datasets. 

Furthermore, by investigating LLMs' implicit understanding to the input with the prober, this work also deepens understanding of how LLMs do reasoning with contexts internally.
\end{abstract}

\section{Introduction}
Large language models (LLMs) demonstrate impressive performance in various downstream tasks.
As the scope of their applications expands, the input lengths of LLMs are increasing rapidly in associated scenarios, i.e., long-context settings~\citep{dubey2024llama,yang2024qwen2,liu2024deepseek}.
Although the input context window becomes larger for latest models, existing LLMs might still exhibit significant performance degradation due to being distracted by input noise~\citep{anil2024many, mo2026opendecoder}.

Such distraction is especially obvious in long-context tasks, e.g., retrieval-augmented generation~\cite{lewis2020retrieval} with external evidence and conversational scenarios~\cite{mo2025uniconv} with multi-turn historical interactions.  In these scenarios, only a small fraction of the provided context is directly relevant to the user input query, whereas the remaining information would act as noise. 
The challenging part is that the noise is often contextually similar to the useful information, which potentially hinders the model from focusing on the most relevant content. 
Besides, processing useless context also affects efficiency.

To this end, existing approaches to mitigate this issue fall into two categories. The first involves multi-agent collaboration frameworks~\citep{qwen-agent-2405, lee2024readagent}, where the context is split into segments processed by different agents, followed by inter-agent interaction to produce a final output. 
While effective, these methods often require complex agent interaction designs and multiple inference steps, resulting in increased latency. 
The second approach, known as ``Parallel Context Encoding''~\citep{yen-etal-2024-long, merth2024superposition}, either trains an additional encoder to process context segments before feeding them to the LLM or trains a separate module to aggregate outputs from multiple LLM runs on different segments. However, these solutions require sophisticated training of new components in addition to the LLM itself, which highly rely on the supervision signal and learning paradigms.

A more feasible way is to explore whether it is possible to leverage an LLM's inherent ability to conduct denoising, e.g., processing the noisy among the internal mechanisms.
Inspired by previous studies suggesting that LLMs implicitly know when they are generating incorrect responses~\citep{kadavath2022language, yin-etal-2023-large}, we hypothesize that LLMs can identify whether the input contains useful information related to the questions before generating the first response token. 

We first test this hypothesis by employing a simple linear prober to investigate whether the LLMs can distinguish it by leveraging the hidden representation in very early layers.
As shown in Section~\ref{sec: Hypothesis Test}, this assumption is strongly supported by the results, which demonstrate that changes in early layers of LLMs can indeed be used for noise detection.

On top of these findings, we further propose a novel approach to conduct context denoising for language models based on early layers in the model architecture.
Our method segments the input into multiple chunks and processes them in parallel. Then, a linear prober, operating on hidden states from early layers, distinguishes between useful and noisy chunks. 
This mechanism discards irrelevant context early to prevent noise distraction in LLMs
The remaining chunks are then combined and processed in a full forward pass. 
We evaluate on two widely used QA datasets and a NoisyRetrieval dataset constructed for the hypothesis test.
Compared to strong baselines, including direct LLM noise discrimination approach and RAG, our approach achieves over 10\% performance improvement and reduces computation by approximately 50\%.

Our main contributions include:
\begin{itemize}[left=0pt, itemsep=0pt]
    \item We identify LLMs' noise sensitivity: Even trivial noise distracts LLMs, while noise resembling target information severely degrades performance.
    \item We find that LLMs can internally differentiate relevant and irrelevant context at lower layers, which can be effectively exploited via a linear prober.
    \item We propose \textsc{END}: By selectively processing input chunks, \textsc{END} mitigates noise distraction effectively and efficiently. Comprehensive experiments confirm its superiority over baselines.
\end{itemize}

\section{Hypothesis Test for Noise Detection via Early Layers in Language Models}
\label{sec: Hypothesis Test}
In this section, we first construct a synthetic task to examine how the harmfulness caused by noise in the context and the impact of longer contexts, and then to test the hypothesis of whether the noise in the long context could be detected by using the early layers in language models.

\subsection{NoisyRetrieval: Synthetic Task Construction for Noise Impact Evaluation}
We first construct NoisyRetrieval for our hypothesis detection.
The key idea is to retrieve an item characterized by five attributes, including \textit{NAME}, \textit{MATERIAL}, \textit{COLOR}, \textit{BRAND}, and \textit{ITEM}.
Specifically, for each question, we add a single true segment as a positive sample, i.e., containing the answer, together with 12 distraction segments that range in five noise difficulty levels as negative ones.
These distraction segments have information similar to the true answer, but do not contain the correct answer.
To generate these negative segments, we control the number of shared attributes between the distractor and the positive segment.
Here, \texttt{Level\_4} is the most challenging, as the distractor shares 4 attributes with the positive segment, whereas \texttt{Level\_0} is the easiest, with no shared attributes.

\begin{table}[ht]
\centering
\begin{tabular}{ccc}
\hline
    Difficulty Level         & Standard    &Long\\ 
    \hline
    Level\_0            & 1.00                  & 1.00 \\ 
    Level\_1            & 1.00                  & 0.94 \\ 
    Level\_2            & 0.91                  & 0.71 \\ 
    Level\_3            & 0.54                  & 0.40 \\ 
    Level\_4            & 0.08                  & 0.04 \\ 
    \hline
\end{tabular}
\caption{Comparison of performance~(accuracy) on our constructed NoisyRetrieval dataset between "Standard"~($\approx$ 4k) and "Long"~($\approx$ 8k) length for \texttt{Llama3-Instruct-8B} with different Difficulty Levels, where Level\_0 is the least confusable, and Level\_4 is the most confusable for the language model.} 
\label{tab:noise}
\end{table}

\begin{figure*}[t]
  \centering
  \includegraphics[width=1.0\textwidth]{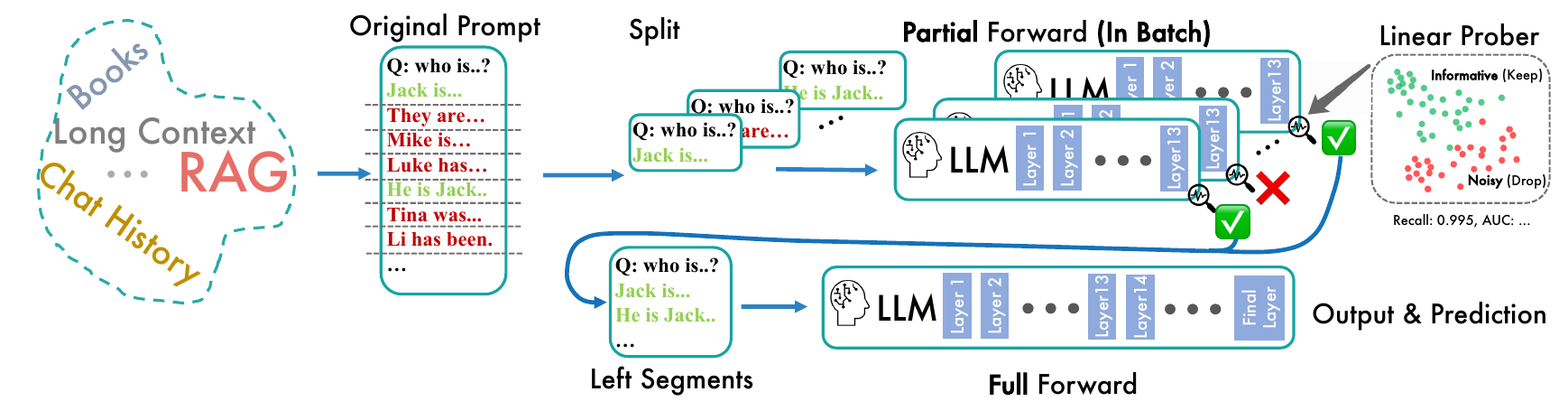}
  \caption{The framework of our proposed approach. The long input, which might come from various sources such as RAG, will be split into chunks for \textit{parallel} processing first. It's a partial forward. A linear prober is attached to the designated layer~(layer 13 in the diagram) of an LLM, and the prober is used to determine whether a chunk is noisy~(\textcolor{red}{red}) or not~(\textcolor{green}{green}). After that, those selected informative chunks will be combined together as the new input for the LLM, which will generate the final prediction.}
  \label{fig:framework}
\end{figure*}

\subsection{The Impact of Noisy Contexts}
In Table~\ref{tab:noise}, we show how the performance of \texttt{Llama3-8B-instruct} model on our synthetic task \textbf{NoisyRetrieval} under two key factors: 1) the difficulty of added noise and 2) overall context length. 
In this experiment, we consider the standard context setting for each instance, which has approximately 4k tokens. Additionally, we test an extended context setting with trivial noise (2× length, $\approx$ 8k tokens) by adding random texts.
We observe that even this seemingly harmless extra material leads to worse performance, particularly when combined with harder noise~(e.g., Level\_4). 
Thus, we obtain two conjectures:
1) The sharp decrease in performance as the noise level increases highlights the critical need for removing noise from the input contexts.
2) The performance gap between the "Standard" and "Long" settings demonstrates that even trivial noise in a longer context can cause severe performance degradation. 
Such results indicate the requirements of reducing input length and mitigating noise to optimize LLM performance.

\begin{figure*}[h]
  \centering
  \includegraphics[width=0.9\textwidth]{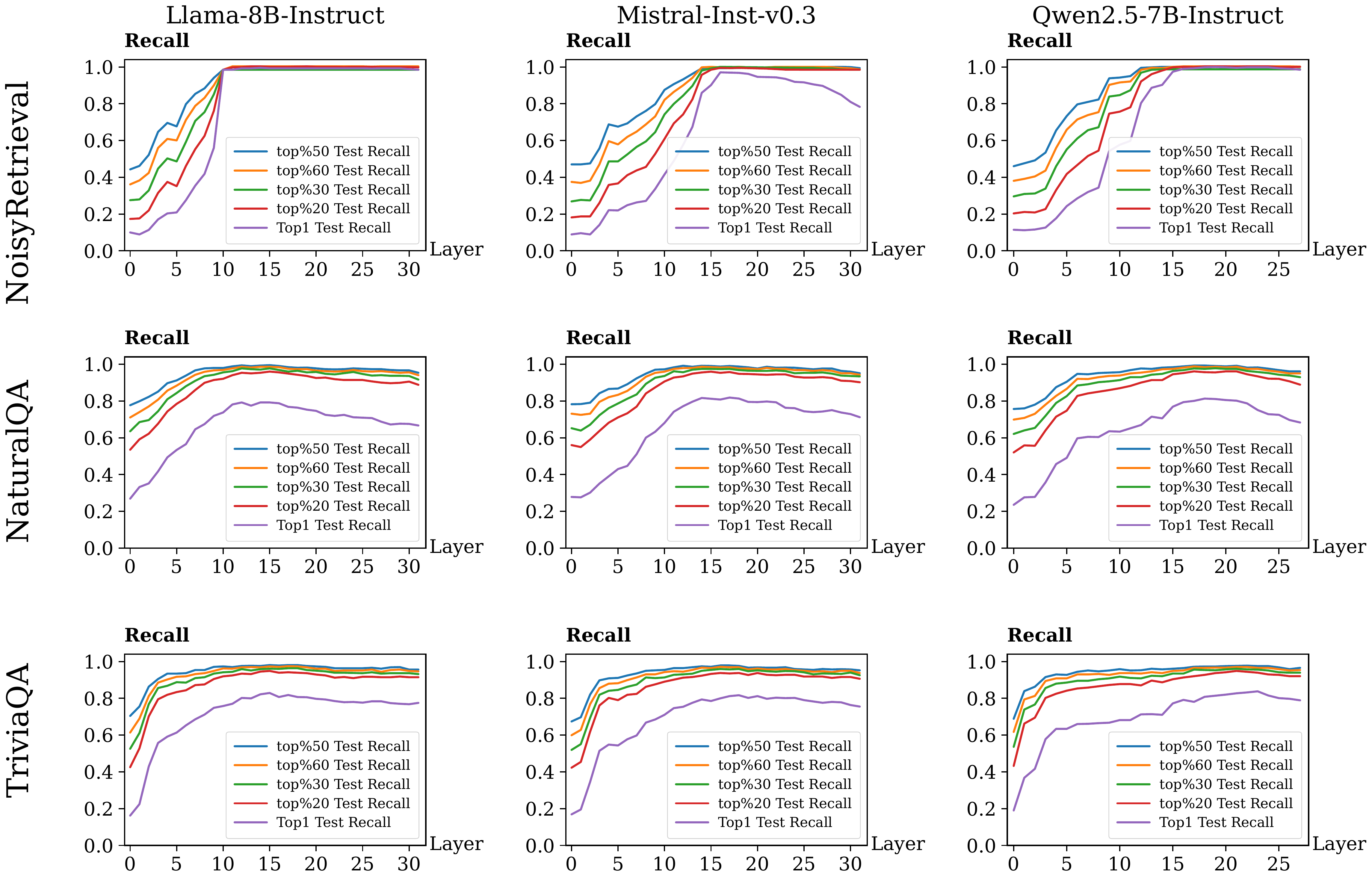}
  \caption{
  This figure shows the recall for positive input of the linear prober when attached to different layers of LLMs. The y-axis represents the recall value, while the x-axis indicates the index of LLM layers. Layer 0 corresponds to the first layer of the model. The top four experiments were conducted on NoisyRetrieval, the middle section on NaturalQA, and the bottom four on TriviaQA. "Top-1 recall" indicates that only the input with the highest score is predicted as positive. "Top 20\% recall" means that inputs with scores in the top 20\% from the linear prober are predicted as positive. Similar interpretations apply for "top 30\% recall", "top 50\% recall", and "top 60\% recall"
}
  \label{fig:prober}
\end{figure*}

\subsection{Hypothesis Test: A Linear Prober can Effectively Elicit LLMs' Inherent Discrimination Capabilities.}
Previous studies~\cite{slobodkin2023curious} on investigating LLMs' implicit abilities suggest that the hidden representation of the last layer in LLMs can be used to reveal whether the generated content is hallucination or not.
Intuitively, we might suppose the LLMs could inherently check whether the input contains information related to the input queries.
To validate this hypothesis, we constructed linear probers for three models: \texttt{Llama3-8B-Instruct}, \texttt{Mistral-Inst-v0.3}, and \texttt{Qwen2.5-7B-Instruct}, across three datasets: \textit{NoisyRetrieval}, \textit{NaturalQA}, and \textit{TriviaQA}.
The functionality of the probers is to predict whether the input is useful.
Following ~\citet{slobodkin2023curious}, the linear prober takes the hidden representation of the last input tokens and predicts binary labels to judge whether the information is useful or not on top of them.
We also conducted a layer-wise analysis to reveal how these inherent abilities emerge in LLMs. 

The results are shown in Figure~\ref{fig:prober}. We observe that LLMs can successfully determine whether the input is noisy via a simple linear prober.
Such discrimination abilities arise at very early layers. For example, using the hidden states from layers around layer 13, the linear prober can already achieve a high recall (> 0.95) for the positive input, i.e., the non-noise input, which confirms our conjectures.
Besides, Different models exhibit similar behavior, suggesting that these inherent abilities are universal across LLMs.

\section{Methodology}
After validating the hypothesis of detecting noise via early layers in language models, we propose an approach to process segmented long input with the help of a linear prober to detect the noise.
The core idea is to segment the input sequence into multiple chunks, process these chunks in parallel through the early layers of the LLM, identify and retain only the essential chunks, and finally forward the retained chunks for the final generation. 
This approach leverages the LLM's inherent ability to discriminate between useful and noisy input early in its processing stages. 
The overall framework is shown in Figure~\ref{fig:framework}, which includes three steps.

\paragraph{Step 1: Input Segmentation.} The first step is to divide the input sequence into multiple chunks. The optimal chunk size may vary depending on the specific task and model architecture. Most chunks will only contain noise rather than the answer in terms of the task. These chunks are then processed in parallel through the early layers of the LLM, allowing for efficient computation. During this parallel processing, each chunk will be attached to a task-related query.

\paragraph{Step 2: Noise Discrimination and Chunk Dropping.}
Then, we leverage LLMs' inherent noise discrimination capability shown in hypothesis test to discriminate noisy chunks and drop them. 
We employ the linear prober described previously as the discriminator, attached to the early layers of the LLM. Specifically, we use a logistic regression model as the prober and attach it to layer 13, based on the observation of hypothesis test (Section~\ref{sec: Hypothesis Test}). 
The prober will predict a score for each chunk, which provide the signals to drop all chunks with a prediction score lower than a specific threshold. 
Note that it is not necessary for the linear prober to predict with 100\% accuracy whether a segment is noisy or not. 
As long as it can filter out most noisy segments, we could obtain more effective and efficient downstream task performance.
Thus, to prevent the information loss and keep the trade-off, we keep chunks with the top 30\% predicted scores.

\paragraph{Step 3: Continue the Forward Pass with Retained Chunks.}
To obtain the final generation after noise reduction with the linear prober, we perform a complete forward pass with the remaining segments. 
To achieve it, we combine all retained chunks and feed them into the LLMs in a new forward pass. 

Considering the parallel processing of the initial inputs, the prober applies at early layers, and the largely reduced input length for the final prediction, our approach does not enforce efficiency burden at the inference time.

\begin{table*}[h!]
\begin{tabular}{lllcc|cc}
\hline
Model & Task & Subset & RAG & LLM-Discrim & \textsc{Ours} & Prober Recall \\
\hline
\multirow{8}{*}{Llama3-8b-instruct} & \multirow{5}{*}{NoisyRetrieval} & Level\_0 & 1.0 & 1.0 & \textbf{1.0} & 1.0\\
 &  & Level\_1 & 1.0 & 1.0 & \textbf{1.0} & 1.0 \\
 &  & Level\_2 & 0.95 & 1.0 & \textbf{1.0} & 1.0 \\
 &  & Level\_3 & 0.51 & 0.98 & \textbf{0.98} & 1.0 \\
 &  & Level\_4 & 0.07 & 0.20 & \textbf{0.68} & 0.99 \\
\cline{2-7}
 & \multirow{2}{*}{Natural QA} & Easy & 69.4 & \textbf{70.69} & 69.8 & 1.0 \\
 &  & Hard & 58.18 & 58.72 & \textbf{61.47} & 0.97\\
\cline{2-7}
 & Trivia QA & Hard & \textbf{37.83} & 37.6 & 37.34 & 0.96 \\
\hline
\multirow{8}{*}{Qwen2.5-7b-instruct} & \multirow{5}{*}{NoisyRetrieval} & Level\_0 & 1.0 & 0.96 & \textbf{1.0} & 1.0 \\
 &  & Level\_1 & 1.0 & 0.96 & \textbf{1.0} & 1.0 \\
 &  & Level\_2 & 1.0 & 0.96 & \textbf{1.0} & 1.0 \\
 &  & Level\_3 & 0.81 & 0.96 & \textbf{0.98} & 1.0 \\
 &  & Level\_4 & 0.52 & \textbf{0.89} & 0.79 & 0.99 \\
\cline{2-7}
 & \multirow{2}{*}{Natural QA} & Easy & \textbf{70.18} & 60.09 & 67.8 & 0.96 \\
 &  & Hard & 55.37 & 54.70 & \textbf{59.07} & 0.95 \\
\cline{2-7}
 & Trivia QA & Hard & 35.71 & 36.40 & \textbf{36.93} & 0.92 \\
\hline
\multirow{8}{*}{Mistral-v0.3-instruct} & \multirow{5}{*}{NoisyRetrieval} & Level\_0 & 1.0 & 0.93 & \textbf{1.0} & 1.0 \\
 &  & Level\_1 & 1.0 & 0.93 & \textbf{1.0} & 1.0 \\
 &  & Level\_2 & 0.96 & 0.93 & \textbf{1.0} & 1.0\\
 &  & Level\_3 & 0.64 & 0.93 & \textbf{0.99} & 1.0 \\
 &  & Level\_4 & 0.25 & \textbf{0.90} &  0.78 & 0.99 \\
\cline{2-7}
 & \multirow{2}{*}{Natural QA} & Easy & \textbf{61.76} & 48.39 & 57.53 & 1.0 \\
 &  & Hard & \textbf{48.58} & 45.66 & 47.88 & 0.94 \\
\cline{2-7}
 & Trivia QA & Hard & 35.63 & \textbf{37.32} & 37.23 & 0.95 \\
\hline
\end{tabular}
\caption{The performance of our proposed approach and the compared baselines. For NoisyRetrieval, we use exact match accuracy as the metric. For NaturalQA and TriviaQA, we use F1 as the metric. We also listed the performance for the linear prober in the Chunk Dropping stage (Prober Recall). }
\label{tab:results}
\end{table*}

\section{Experiments}

\subsection{Experimental Setup}
\paragraph{Dataset.}
\label{sec:dataset}
We use three datasets to evaluate our proposed methods, including two widely used QA datasets, Natural QA~\cite{kwiatkowski2019natural} and Trivia QA~\cite{joshi2017triviaqa}, and a self-constructed one, NoisyRetrieval, for the noise detection experiment.

\noindent\textbf{Natural QA} is a QA dataset that focuses on factoid questions, where each instance has positive and negative segments retrieved in advance. 
Each instance has negative segments defined as two noise levels, `Easy' and `Hard', according to the similarity score. Hard negative segments might distract LLMs more easily because they have higher similarity scores in terms of the input query. 

\noindent\textbf{Trivia QA} is similar to Natural QA, where each instance has positive and negative segments provided in advance. Each instance has the negative segments with one noise level as ‘Hard’.

\noindent\textbf{NoisyRetrieval} is constructed to require retrieving the prober of a long context input characterized by five attributes. Each instance contains a positive segment and 12 noisy negative segments that serve as distractors. 
More details about this task and concrete examples can be found in Appendix~\ref{apdx:needle}.

\paragraph{Baselines.}

To ensure a fair comparison across scenarios with varying noise levels and context lengths, we establish two competitive baselines: (1) \textbf{RAG}~\cite{lewis2020retrieval}, which simulates a full pipeline. To implement it, we combine positive segments with negative segments as the post-reranking inputs for generation; (2) \textbf{LLM-Discrim}~\citep{zheng2024judging, fu2023gptscore} is a state-of-the-art baseline, which has two forward passes. In the first forward pass, it splits the input into chunks and asks the LLMs to determine whether a segment contains the answer. Then, in the second forward pass, it requires the LLMs to provide the prediction using the remaining chunks as input.

\subsection{Main Results}

\noindent\paragraph{Effectiveness.} The main results are reported in Table~\ref{tab:results}. We can see that our proposed approach significantly outperforms the RAG baseline, and it also achieves the best overall performance even when compared to the strong LLM-Discrim baseline. 
Besides, the superiority of our approaches becomes more pronounced on more challenging tasks, i.e., greater difficulty input.
As expected, the recall of the linear prober is sufficiently high. 

One potential disadvantage is using a single metric for evaluation.
For example, in the case of \textit{NaturalQA (Easy)} with \texttt{Mistral-Inst-v0.3}, the recall is nearly perfect (0.998), yet the F1 score is still lower than anticipated. 
This discrepancy suggests that while our method is highly effective at identifying relevant information, the overall performance metric could be further expanded to fully capture the systems' advantage in certain scenarios.

\begin{table}[t]
\centering
\small 
\renewcommand{\arraystretch}{1.8} 
\begin{tabular}{lll}
\hline
\textbf{Method} & \textbf{Cost Expression} & \textbf{Complexity} \\ \hline
RAG & $\Theta((10L)^2)$ & $\Theta(100L^2)$ \\
LLM-Discrim & $\Theta((10L)^2 + (2L)^2)$ & $\Theta(104L^2)$ \\
\textsc{Ours} & $\Theta\left(\frac{13}{32}(10L)^2 + (3L)^2\right)$ & $\Theta(49\frac{5}{8}L^2)$ \\ \hline
\end{tabular}
\caption{Comparison of computational costs between baselines and our proposed method.}
\label{tab:computational_cost}
\end{table}

\noindent\paragraph{Efficiency.}
While our approach appears to require an additional forward pass due to the extra probing and Chunk Dropping, it actually still processes a significantly reduced input during its second forward pass compared to the \texttt{RAG} baseline, owing to the substantial chunk elimination. 
Moreover, in comparison with the strong \texttt{LLM-Discrim} baseline, the Chunk Dropping stage consumes less than half of a forward pass. 
To estimate the computational cost, we provide an accurate efficiency comparison as described in Table~\ref{tab:computational_cost}.
This is based on the assumption of quadratic complexity for LLM operations, and considering an input segmented into 10 chunks, each of length $L$, and dropping 70\% of chunks at layer-13 based on \texttt{Llama-3-8B-Instruct}~(32 layers).
The approximation procedure is shown below:
\texttt{RAG}: $\Theta((10L)^2) = \Theta(100L^2)$;
\texttt{LLM-Discrim} (assuming 2 chunks retained): $\Theta((10L)^2 + (2L)^2) = \Theta(104L^2)$;
\texttt{\textsc{Ours}} (30\% chunks retained): $\Theta(\frac{13}{32}(10L)^2 + (3L)^2) = \Theta(49\frac{5}{8}L^2)$.
Such an estimation demonstrates that our approach achieves superior efficiency compared to both baselines. Real-world wall-clock time analysis also reveals the efficiency advantage as shown in Appendix~\ref{apdx:efficiency}.

\subsection{Analysis of the Linear Prober}
The efficacy of our proposed approach is heavily influenced by the performance of the linear prober. A more accurate and precise linear prober results in shorter final inputs, enabling LLMs to generate responses more efficiently with a better understanding of the contexts. 
This section presents a comprehensive analysis of the linear prober, aiming to demonstrate that it unveils LLMs' inherent capabilities and implicit reasoning processes when addressing factual queries based on the input.

\begin{table}
    \centering
    \small
    \setlength{\tabcolsep}{7pt}
    \scalebox{0.9}{
    \begin{tabular}{cccc}
    \hline
        Backbone & NoisyRetrieval & NaturalQA & TriviaQA \\
    \hline
        Llama3-rand & 0.60 & 0.71 &  0.66 \\ 
        Llama3 & 0.99 & 0.91 & 0.90 \\
    \hline
        Qwen2.5-rand & 0.64 & 0.71 & 0.67 \\ 
        Qwen2.5 & 0.95 & 0.88 & 0.86 \\
    \hline
        Mistral-rand & 0.60 & 0.70 & 0.66 \\ 
        Mistral & 0.93  & 0.89 & 0.89 \\
    \hline
    \end{tabular}}
    \caption{Comparison of linear prober with F1 scores between pre-trained and randomly initialized models (with -rand suffix).} 
    \label{tab:random}
\end{table}

\paragraph{Linear Prober's Effectiveness is Different from Fitting.} 
To verify that the linear prober's discriminative capability does not stem from the fitting process of training the prober itself, we conducted experiments using randomly initialized LLMs as the backbone while maintaining the original embedding layer to preserve semantic representation. The performance comparison between these randomly initialized LLMs and their pre-trained counterparts is presented in Table~\ref{tab:random}. 
The substantial performance gap of the linear prober demonstrates that this implicit discriminative ability originates from the pre-trained LLMs rather than training the linear prober with a LLM as a general feature extractor. 

\begin{table}
    \centering
    \small
    \setlength{\tabcolsep}{7pt}
    \scalebox{0.95}{
    \begin{tabular}{lccc|r}
    \hline
        Task & [S|C|Q] & [S|Q|C] & [Q|S|C] & [S|C] \\
    \hline
        NoisyRetrieval & 1.0 & 0.98 & 0.96 & 0.25 \\ 
        NaturalQA & 0.97 & 0.83 & 0.84 & 0.36 \\
        TriviaQA & 0.96 & 0.70 & 0.68 & 0.36 \\
    \hline
    \end{tabular}}
    \caption{The top 30\% recall of the linear prober (layer-15) across different prompt formats.}
    \label{tab:prompt}
\end{table}

\paragraph{The Linear Prober is Robust to Prompt Variations.} Previous studies has shown that LLMs can be sensitive to prompt formats~\citep{sclar2023quantifying}. To explore the robustness of the linear prober, we analyze its performance across four different prompt formats: [S|C|Q], [S|Q|C],[Q|S|C], and [S|C], where `S', `Q', and `C' represents the system prompt, the question, and the context, respectively. 
As shown in Table~\ref{tab:prompt}, although the linear prober is only trained on [S|C|Q], it generalizes well across [S|C|Q], [S|Q|C] and [Q|S|C] formats. Such kind of generalization  strongly supporting the hypothesis that LLMs are inherently prepared to answer a question once the context is fully presented. 

In contrast, the performance degradation on the [S|C] format, which suggests that the linear prober relies on more than just distinguishing between noise and informative input. This might because it extracts the model's implicit understanding of the input in relation to the question. The specific prompts used for each dataset and format are provided in Appendix~\ref{apdx:prompt}.


\begin{figure}[t]
    \centering
    \includegraphics[width=1.0\linewidth]{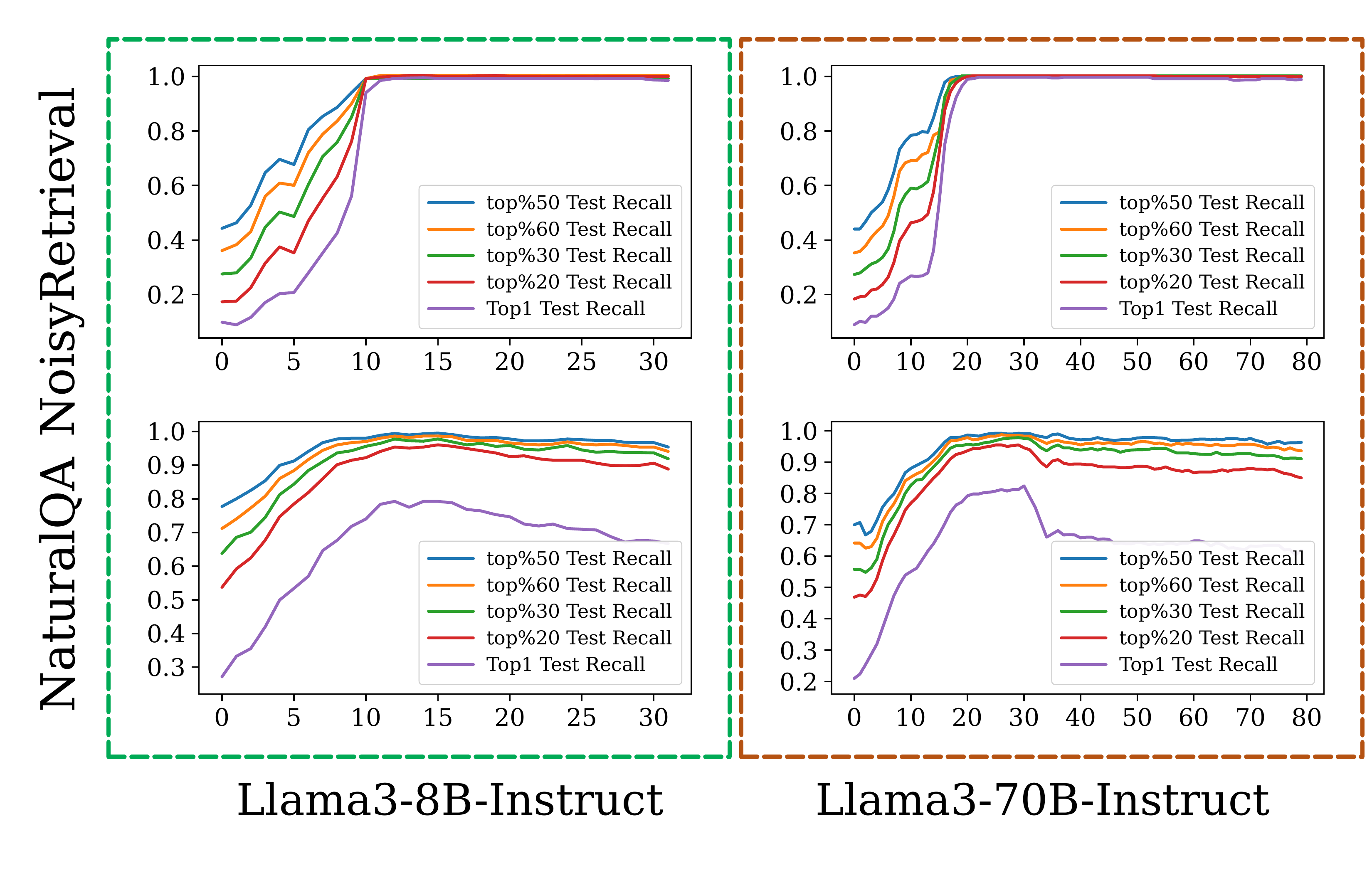}
    \caption{The performance for the linear prober with \texttt{Llama-3-8B-Instruct} and \texttt{Llama-3-70B-Instruct} as the backbone.}
    \label{fig:large_small}
    \vspace{-10pt}
\end{figure}

\paragraph{The Distinguishing Capabilities between Large and Small Models Exhibit at Similar Layers.} We conduct probing experiments on both Llama3-70B-Instruct and Llama3-8B-Instruct models to investigate how their distinguishing abilities evolve across layers. The performance is shown in Figure~\ref{fig:large_small}.
Despite the significant difference in size, i.e., Llama3-70B-Instruct has 80 layers, while Llama3-8B-Instruct has only 32 layers, the performance of the linear prober saturates around similar layers for both models, specifically between layers 10 and 15. This finding suggests that the key distinguishing capabilities in language models is not solely dependent on model size but is instead strongly tied to the depth. Even in the larger Llama3-70B-Instruct model, these capabilities arise early in the network, around the same layer range as in the much smaller Llama3-8B-Instruct. This consistency across models of varying sizes implies that these distinguishing features are fundamental properties of the model's internal representations and layer-wise progression.


\begin{figure}[t]
    \centering
    \includegraphics[width=1.0\linewidth]{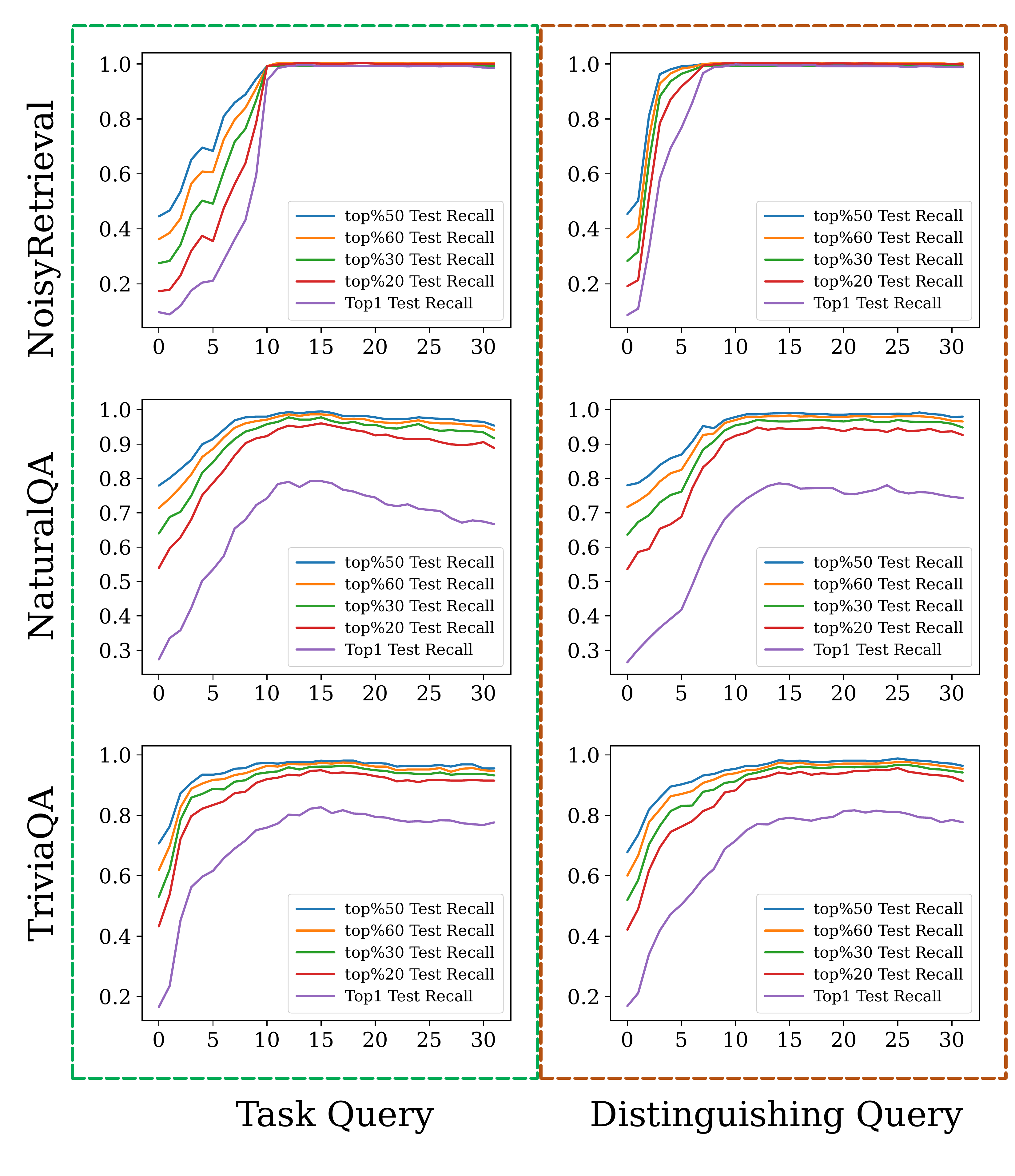}
    \caption{The performance for the linear prober with \texttt{Llama-3-8B-Instruct} as the backbone. The left three figures use the task query in the prompts. The right three figures use a distinguishing query in prompts.}
    \label{fig:query_cmp}
    \vspace{-10pt}
\end{figure}

\paragraph{The Implicit Discrimination Ability is as Strong as Explicitly Requiring LLMs to Discriminate the Input.}
We conducted an experiment where we replaced the task query in the prompt with a distinguishing query that directly asks the LLM to determine whether the input is relevant to the question, similar to the first forward pass of the baseline method \textbf{LLM-Discrim}. 
We then attached the linear prober to the representations generated from this input to evaluate whether this explicit requirement would lead to improved performance. 
The details of distinguishing query are provided in Appendix~\ref{apdx:distinguishing}.
The results are shown in Figure~\ref{fig:query_cmp}. While the performance with the distinguishing query appears slightly better than with the task query, the overall performance of the linear prober remains nearly the same. 
This outcome supports our earlier hypothesis again: regardless of whether the model is explicitly instructed to distinguish relevant input, LLMs seem to perform this discrimination step implicitly before answering questions related to context.

\section{Related Work}

\noindent\paragraph{Enhancing LLMs' Long Context Ability.}
\citet{zhang2024found} found that positional encoding decay weakens self-attention on relevant parts and proposed modifying key attention head dimensions to improve LLMs' performance on noisy input. \citet{hsieh2024found} addressed positional bias by introducing a calibration mechanism that subtracts baseline attention from a dummy document, preventing overemphasis on input boundaries. \citet{he-etal-2024-never} tackled this via instruction tuning to help LLMs focus on target information. Beyond data-centric approaches, \citet{wu2024reducing} applied contrastive loss in post-training, enhancing robustness by retrieving similar document pairs.

\noindent\paragraph{Retrieval Augmented Generation (RAG).} Different from directly feeding all collected texts to query the LLMs, RAG first performs chunking and then retrieves the most related chunks across all candidate texts~\citep{lewis2020retrieval,asai2023self,zhang2025ratt}. Although RAG can significantly boost the performance of LLMs, it still suffers from the distraction problem as we will show below. This is because there are still only a few useful chunks among all the retrieved chunks, where the noisy ones could degrade the performance~\cite{su2025parametric}.

\noindent\paragraph{Prompt/Context Compression.}
The original prompts always contain many useless tokens, and this line of works automatically prunes redundant tokens in the prompts so as to reduce the input length. They leverage tailored metrics such as self-information to keep important information~\citep{li2023compressing}. Together, these studies demonstrate that strategic compression of input contexts can lead to computational savings without sacrificing accuracy~\citep{jiang2023llmlingua, xu2023recomp,liu2025smooth}.

\noindent\paragraph{Separate Context Processing.}
This approach decomposes long inputs into separate parts for individual processing, including agent collaboration and parallel context processing. Multi-agent systems iteratively exchange results to refine predictions~\citep{zhao2024longagent, qwen-agent-2405, lee2024readagent}. Parallel processing extracts and merges useful information~\citep{yen-etal-2024-long} or employs trainable selection mechanisms to determine predictions~\citep{merth2024superposition}. Some methods apply KV-cache compression post-segmentation to reduce noise and context length~\citep{kim2024infinipot}.

\noindent\paragraph{Linear Probing in Natural Language Processing.}
Linear probing has been used to extract knowledge from models, including LLMs~\citep{gurnee2023language}. Recent NLP studies show that applying it to the first generated token can aid trust-related tasks like hallucination detection~\citep{slobodkin2023curious}. Unlike prior work, we perform layer-wise probing and find that LLMs assess input informativeness at an early stage. Moreover, while existing methods probe task-related concepts, our approach explores unrelated concepts.

\section{Conclusion}

In this paper, we introduce a context denoising framework for language models based on their early architecture layers, which aims to improve LLMs performance when processing noisy or irrelevant context. Our approach effectively identifies and removes noisy input chunks early in the LLM pipeline, improving performance across various tasks. 
This method demonstrates versatility by working well across different LLM architectures without requiring fine-tuning. By reducing unnecessary computation on irrelevant information, our approach offers both performance and efficiency gains.

While our approach shows significant promise, future work could explore more advanced chunking strategies and dynamic thresholding techniques. The prober could be enhanced with additional trainable parameters or greater complexity.

\section{Limitations}

The generalization of linear prober across tasks remains challenging as our analysis provided in Appendix~\ref{apdx:generalization}. While fitting a simple linear regression requires minimal data, this limitation could potentially be addressed through the implementation of a more sophisticated probing mechanism. 
Besides, another potential risk is that the mechanism of LLMs' internal behavior is still not entirely clear, requiring further investigation. 
In addition, due to our limited computational resources, we did not conduct extensive experiments on large LLMs such as Llama-3-70B. Nevertheless, the prober still performs well at this scale and appears to save more computation as the results shown in Appendix~\ref{apdx:large_prober}. 

\bibliography{custom}


\clearpage

\appendix

\section*{Appendix}

\section{Dataset Statistics}
The detailed statistics of the dataset are as follows. After filtering out instances without a sufficient number of negative chunks, the dataset used in our study has the following distribution:

\begin{table}[h!]
    \centering
    \small
    \begin{tabular}{lccc}
    \toprule
    \textbf{Dataset} & \textbf{Train Set (Prober)} & \textbf{Test Set} \\
    \midrule
    NaturalQA & 1000 & 2653 \\
    TriviaQA & 1000 & 2619 \\
    \makecell{NoisyRetrieval \\ (each difficulty level)} & 1000 & 2000 \\
    \bottomrule
    \end{tabular}
    \caption{Dataset statistics}
    \label{tab:dataset_statistics}
\end{table}

\section{Prober Training Details}
\label{apdx:prober} 
The prober was trained using the following steps: 
\begin{enumerate}
\item For each question (instance), collect negative-positive segment pairs. 
\item Label negative segments as 0 and positive segments as 1. 
\item Feed these segments, along with their corresponding questions, into the model to extract intermediate representations. 
\item Split all instances, along with their corresponding negative and positive segments, into training and test sets. 
\item Train a simple sigmoid-based linear prober (logistic regression). 
\end{enumerate} 
The training set for each prober on each dataset consists of 1,000 instances. Further details about these data can be found in the previous section.

\begin{table*}[ht]
\centering
\begin{tabular}{lccccc}
\toprule
Model/Method          & Chunk Number & 4k       & 8k       & 16k     & 32k     \\
\midrule
\textbf{Llama3-8B-Ours}  & \textbf{10}  & \textbf{1.4810s} & \textbf{1.315s} & \textbf{1.887s} & \textbf{3.109s} \\
\textbf{Llama3-8B-Ours}  & \textbf{20}  & \textbf{1.508s}  & \textbf{1.314s} & \textbf{1.871s} & \textbf{3.052s} \\
Llama3-8B            & NA           & 1.226s   & 1.846s   & 2.996s   & 6.173s \\
\textbf{Llama3-70B-Ours} & \textbf{10}  & \textbf{5.049s} & \textbf{6.069s} & \textbf{OOM}  & \textbf{OOM} \\
Llama3-70B           & NA           & 6.416s   & 9.998s   & OOM      & OOM \\
\bottomrule
\end{tabular}
\caption{Wall-clock inference times (in seconds) with the flash-attention kernel for different models/methods and sequence lengths. \textbf{xxx-Ours} indicates models equipped with the proposed method; ``Chunk Number'' denotes the number of segments for each input.}
\label{apdx_tab:inference_times}
\end{table*}

\section{More results of the Prober's performance}
\subsection{Generalization of The Linear Prober}
\begin{figure}[t]
    \centering
    \includegraphics[width=1.0\linewidth]{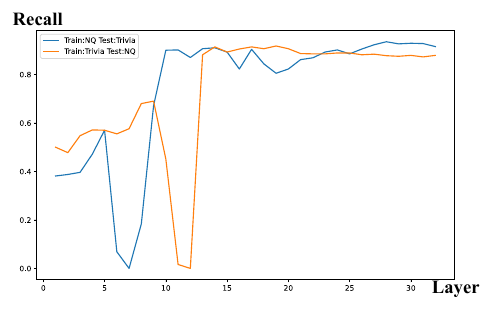}
    \caption{The cross-task performance for the linear prober with \texttt{Llama-3-8B-Instruct} as the backbone.}
    \label{apdx_fig:cross_recall}
\end{figure}

We tested cross-task generalization with two settings: The prober is trained with NaturalQA, but tested on Trivia; The prober is trained with Trivia, but tested on NaturalQA. As shown in Figure~\ref{apdx_fig:cross_recall}, we found that: Cross-task performance is decent (about 0.9 recall), but not perfect (1.0 recall). It still shows the existing ability of implicit discrimination. Meanwhile, the performance shows some instability.
\label{apdx:generalization}

\subsection{Prober on Larger LLMs}
\label{apdx:large_prober}
\begin{figure}[t]
    \centering
    \includegraphics[width=1.0\linewidth]{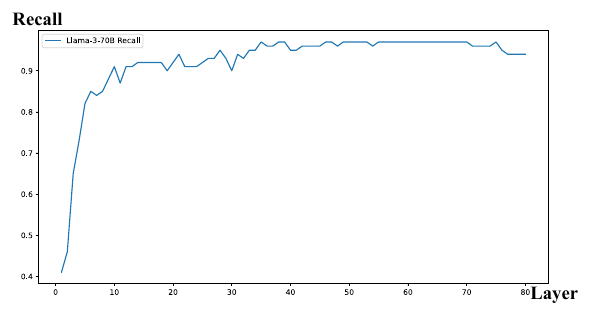}
    \caption{The performance for the linear prober with \texttt{Llama-3-70B-Instruct} as the backbone.}
    \label{apdx_fig:large_recall}
\end{figure}

We conducted probing experiments with Llama-3-70B, which has 80 layers~(Figure~\ref{apdx_fig:large_recall}). The results show that the optimal probing layer is similar to that of much smaller models (e.g., the 8B model), typically around layers 10–15. This finding suggests that our approach remains beneficial for larger models, as the layer-wise performance does not shift significantly and we just need to run the a few layers. The portion of the first partial forward in the whole cost becomes lower (13/32 --> 13/80). However, if \textsc{LLMDiscrim} is used instead, the cost of full forward passes would increase considerably due to the greater number of layers and parameters per layer.

\section{Real-World Efficiency Analysis}
\label{apdx:efficiency}

In real-world LLM deployment, flash-attention~\cite{dao2023flashattention2}, which significantly reduces the quadratic time complexity of self-attention, is widely used. In this section, we demonstrate that our method continues to yield efficiency benefits with the flash-attention kernel. Table~\ref{apdx_tab:inference_times} presents the wall-clock inference times under the same experimental settings as our other experiments: 30\% of the chunks are retained for the second forward pass, and the first forward pass is executed in batch across layers 1 to 13.

Our results indicate that when the input is long (i.e., when the computation cost is dominated by the input length) or when the model is large, our proposed method clearly outperforms the standard approach of processing the entire sequence in a single forward pass. Although the observed efficiency improvement does not precisely match the predictions from our complexity analysis, these findings nonetheless highlight the practical benefits of our method in real-world scenarios.

\section{NosiyNeedle}
\label{apdx:needle}
\subsection*{NoisyRetrieval (Synthetic)}

The original ``Needle-in-a-Haystack'' style task~\citep{hsieh2024ruler} such as passkey retrieval~\citep{mohtashami2023landmark} is too easy. The passkey retrieval task is to retrieve a simple passkey like "The passkey is 12345".
Its contexts are some meaningless texts such as repeated "The grass is green, the sky is blue". An example looks like:
\begin{lstlisting}
The grass is green, the sky is blue. The grass is green, the sky is blue. The passkey is 12345. The grass is green, the sky is blue. The passkey is 12345. The grass is green, the sky is blue. The grass is green, the sky is blue. The grass is green, the sky is blue. What is the passkey?
\end{lstlisting}
This is far from real cases, and any retriever can perfectly find the answer segments. Hence, we design the \textbf{NoisyRetrieval} task as a noisier version. We set five noise levels (level\_0 - level\_4) for segments, by controlling five attributes in the target sentence: \texttt{[Name]}, \texttt{[Material]}, \texttt{[Color]}, \texttt{[Brand]}, \texttt{[Item]}. For example:
\begin{quote}
[Jack's] Password to his [Green] [Wooden] [Benz] [phone] is 34512:
\end{quote}

With this sentence as the answerm, some examples are:
\begin{itemize}
    \item \textbf{level\_0}: Random but meaningful texts from some papers.
    \item \textbf{level\_1}: Random but meaningful texts from some papers + a sentence with one attribute the same as the target:
    \begin{quote}
    [Paul's] Password to his [red] [golden] [Apple] [\textbf{phone}] is 51233
    \end{quote}
    \item \textbf{level\_4}: Random but meaningful texts from some papers + a sentence with four attributes the same as the target:
    \begin{quote}
    [\textbf{Jack's}] Password to his [\textbf{Green}] [\textbf{Wooden}] [\textbf{Benz}] [laptop] is 12345
    \end{quote}
\end{itemize}

An example input for \textbf{NoisyRetrieval}:
\begin{lstlisting}
[other contexts] [Jack's Password to his Green Wooden Apple phone is 12345] [other contexts] [Paul's Password to his Green Wooden Apple phone] [other contexts] [Jack's Password to his red glass Benz phone is 11145] [other contexts] [Jack's Password to his yellow glass Apple phone is 32525] [Noise] ... 
Question: What is Jack's Password to his Green Wooden Benz phone?
Answer:
\end{lstlisting}

For this task, each positive chunks contains the answer inserted randomly into some random texts from random essays. Each negative chunk include distractor content inserted randomly into random texts from random essays. For input context construction, to ensure sufficient difficulty and realism, the chunk containing the answer is placed in the middle of the context. For example, with 10 negative chunks and 1 positive chunk, the positive chunk is positioned as the 6th chunk within the context.

\subsection{Attribute Value}
The possible values of all attributes are listed below:
\begin{lstlisting}
Name = ["John", "Emma", "Alex", "Sophia", "Michael", "Olivia", "Liam", "Ava", "Noah", "Isabella", "Ethan", "Mia", "Mason", "Charlotte", "William", "Amelia", "James", "Harper", "Benjamin", "Evelyn"] 
Color = ["red", "blue", "green", "yellow", "black", "white", "purple", "orange", "pink", "brown", "gray", "navy", "teal", "maroon", "olive", "silver", "gold", "turquoise", "lavender", "coral"]
Item = ["bag", "watch", "phone", "laptop", "headphones", "sunglasses", "shoes", "jacket", "camera", "tablet", "wallet", "backpack", "earbuds", "smartwatch", "keyboard", "mouse", "speaker", "charger", "fitness tracker", "power bank"]
Brand = ["Apple", "Samsung", "Nike", "Adidas", "Sony", "Gucci", "Microsoft", "Dell", "LG", "Bose", "Lenovo", "Asus", "Logitech", "Prada", "Canon", "Nikon", "Fitbit", "Fossil", "JBL", "Anker"]
Material = ["leather", "aluminum", "plastic", "glass", "titanium", "silicone", "ceramic", "fabric","wood", "rubber", "nylon", "polyester", "cotton", "wool", "denim", "suede", "velvet", "cork"]
\end{lstlisting}

\section{Used Prompts}
\label{apdx:prompt}
The prompts used in the probing stage are listed below. [S|C|Q] is the default prompt format without specifying.
\subsection{NaturalQA}
\textbf{[S|C|Q] Setting:}
\begin{lstlisting}
System Prompt: "You are given some pieces of a story, which can be either a novel or a movie script, and a question. Answer the question as concisely as you can, using a single phrase if possible. Do not provide any explanation."

Prompt Template:
"{SYSTEM}
Pieces of the story:
{CONTEXT}
Now, if you can find the required information, answer the question based on the story as concisely as you can, using a single phrase if possible. Do not provide any explanation.

Question: {QUERY}

Answer:"
\end{lstlisting}

\textbf{[S|Q|C] Setting:}
\begin{lstlisting}
System Prompt: "You are given some pieces of a story, which can be either a novel or a movie script, and a question. Answer the question as concisely as you can, using a single phrase if possible. Do not provide any explanation."

Prompt Template:
"{SYSTEM}

Now, if you can find the required information, answer the question based on the story as concisely as you can, using a single phrase if possible. Do not provide any explanation. Question: {QUERY}

Pieces of the story:
{CTX}

Answer:"
\end{lstlisting}

\textbf{[Q|S|C] Setting:}
\begin{lstlisting}
System Prompt: "You are given some pieces of a story, which can be either a novel or a movie script, and a question. Answer the question as concisely as you can, using a single phrase if possible. Do not provide any explanation."

Prompt Template:
"Now, if you can find the required information, answer the question based on the story as concisely as you can, using a single phrase if possible. Do not provide any explanation.

Question: {QUERY}

{SYSTEM}

Pieces of the story:
{CTX}

Answer:"
\end{lstlisting}

\textbf{[S|C] Setting:}
\begin{lstlisting}
System Prompt: "You are given some pieces of a story, which can be either a novel or a movie script, and a question. Answer the question as concisely as you can, using a single phrase if possible. Do not provide any explanation."

Prompt Template:
"
{SYSTEM}

Pieces of the story:
{CTX}

Answer:"
\end{lstlisting}

\subsection{NoisyNeedle}

\textbf{[S|C|Q] Setting:}
\begin{lstlisting}
System Prompt: "There are information about a passkey hidden in input. Please remember it."

Prompt Template:
"{SYSTEM}
Part of the input:
{CTX}
Remeber your task: according to all previous input, if there is the answer to: {QUERY}, answer the question
"
\end{lstlisting}

\textbf{[S|Q|C] Setting:}
\begin{lstlisting}
System Prompt: "There are information about a passkey hidden in input. Please remember it."

Prompt Template:
"{SYSTEM}
Remeber your task: according to all previous input, if there is the answer to: {QUERY}, answer the question
Part of the input:
{CTX}
"
\end{lstlisting}

\textbf{[Q|S|C] Setting:}
\begin{lstlisting}
System Prompt: "There are information about a passkey hidden in input. Please remember it."

Prompt Template:
"Remeber your task: according to all previous input, if there is the answer to: {QUERY}, answer the question
{SYSTEM}
Part of the input:
{CTX}
"
\end{lstlisting}

\textbf{[S|C] Setting:}
\begin{lstlisting}
System Prompt: "There are information about a passkey hidden in input. Please remember it."

Prompt Template:
"{SYSTEM}
Part of the input:
{CTX}

" 
\end{lstlisting}

\subsection{Trivia}

\textbf{[S|C|Q] Setting:}
\begin{lstlisting}
System Prompt: "Answer the question based on the given passage. Only give me the answer and do not output any other words."

Prompt Template:
"{SYSTEM}

The following are some passages:
{CTX}
Now,  if you can find the required information, answer the question based on those passages. Do not provide any explanation.

Question: {QUERY}

Answer:"
\end{lstlisting}

\textbf{[S|Q|C] Setting:}
\begin{lstlisting}
System Prompt: "Answer the question based on the given passage. Only give me the answer and do not output any other words."

Prompt Template:
"{SYSTEM}
Now, if you can find the required information, answer the question based on those passages. Do not provide any explanation.

Question: {QUERY}

The following are some passages: 
{CTX}

Answer:"
\end{lstlisting}

\textbf{[Q|S|C] Setting:}
\begin{lstlisting}
System Prompt: "Answer the question based on the given passage. Only give me the answer and do not output any other words."

Prompt Template:
"Now,  if you can find the required information, answer the question based on those passages. Do not provide any explanation.

Question: {QUERY}

{SYSTEM}

The following are some passages:
{CTX}

Answer:"
\end{lstlisting}

\textbf{[S|C] Setting:}
\begin{lstlisting}
System Prompt: "Answer the question based on the given passage. Only give me the answer and do not output any other words."

Prompt Template:
"{SYSTEM}

The following are some passages:
{CTX}

Answer:"
\end{lstlisting}

\subsection{Distinguishing Prompt}\label{apdx:distinguishing}
The distinguishing prompt directly requires the LLM to discriminate whether the context contains answers to a given question. 

\begin{lstlisting}
Prompt Template:
"
You are a helpful assistant. I will give you a query and some contexts. Please help me with my question about the given query and contexts. 

Question: {QUERY}

The following are some passages:
{CTX}

Do the given contexts contain the answers to the given question? Use 'Yes' or 'No' to answer it. Do not provide any explanation.

Answer:"
\end{lstlisting}

\end{document}